\documentclass{llncs}

\usepackage{amsmath}
\usepackage{graphicx}        % standard LaTeX graphics tool when including figure files
\usepackage[bottom]{footmisc}% places footnotes at page bottom
\usepackage{epstopdf}
\usepackage{color}

\begin{document} 

\title{A Genetic Algorithm for Power-Aware Virtual Machine Allocation in Private Cloud}

\author{Nguyen Quang-Hung\inst{1}, Pham Dac Nien\inst{2}, Nguyen Hoai Nam\inst{2}, \\Nguyen Huynh Tuong\inst{1}, Nam Thoai\inst{1}}

\institute{Faculty of Computer Science and Engineering\\
Ho Chi Minh City University of Technology  \\
268 Ly Thuong Kiet Street, District 10, Ho Chi Minh City, Vietnam\\ 
\inst{1}\textit{\{hungnq2, htnguyen, nam\}@cse.hcmut.edu.vn} \\
\inst{2}\textit{\{50801500, 50801308\}@stu.hcmut.edu.vn}
}

\maketitle

\abstract{
Energy efficiency has become an important measurement of scheduling algorithm for private cloud. The challenge is trade-off between minimizing of energy consumption and satisfying Quality of Service (QoS) (e.g. performance or resource availability on time for reservation request). We consider resource needs in context of a private cloud system to provide resources for applications in teaching and researching. In which users request computing resources for laboratory classes at start times and non-interrupted duration in some hours in prior. Many previous works are based on migrating techniques to move online virtual machines (VMs) from low utilization hosts and turn these hosts off to reduce energy consumption. However, the techniques for migration of VMs could not use in our case.
In this paper, a genetic algorithm for power-aware in scheduling of resource allocation (GAPA) has been proposed to solve the static virtual machine allocation problem (SVMAP). Due to limited resources (i.e. memory) for executing simulation, we created a workload that contains a sample of one-day timetable of lab hours in our university. We evaluate the GAPA and a baseline scheduling algorithm (BFD), which sorts list of virtual machines in start time (i.e. earliest start time first) and using best-fit decreasing (i.e. least increased power consumption) algorithm, for solving the same SVMAP. As a result, the GAPA algorithm obtains total energy consumption is lower than the baseline algorithm on simulated experimentation. 

}

\section{Introduction}

Cloud computing \cite{Buyya2009a}, which is popular with pay-as-you-go utility model, is economy driven. Saving operating costs in terms of energy consumption (Watts-Hour) for a cloud system is highly motivated for any cloud providers. Energy-efficient resource management in large-scale datacenter is still challenge \cite{Albers2010}\cite{VonLaszewski2009}\cite{Goiri2010}\cite{Beloglazov2012}. The challenge of energy-efficient scheduling algorithm is trade-off between minimizing of energy consumption and satisfying demand resource needs on time and non-preemptive. Resource requirements depend on the applications and we are interested in virtual computing lab, which is a cloud system to provide resources for teaching and researching.

There are many studies on energy efficient in datacenters. Some studies proposed energy efficient algorithm that are based on processor speed scaling (assumption that CPU technology supports dynamic scaling frequency and voltage (DVFS)) \cite{Albers2010}\cite{VonLaszewski2009}. Some other studies proposed energy efficient by scheduling for VMs in virtualized datacenter \cite{Goiri2010}\cite{Beloglazov2012}. A. Beloglazov et al. \cite{Beloglazov2012} presents the Modified Best-Fit Decreasing (MBFD) algorithm, which is best-fit decreasing heuristic, for power-aware VM allocation and adaptive threshold-based migration algorithms to dynamic consolidation of VM resource partitions. Goiri, Í. et al. \cite{Goiri2010} presents score-based scheduling, which is hill-climbing algorithm, to place each VM onto which physical machine has the maximum score. However, the challenge is still remain. These previous works did not concern on satisfying demand resource needs on time (i.e. VM starts at a specified start time) and non-preemptive, in addition to both MBFD and score-based algorithms do not find an optimal solution for VM allocation problem.

In this paper, we introduce our static virtual machine allocation problem (SVMAP). To solve the SVMAP, we propose the GAPA, which is a genetic algorithm to find an optimal solution for VM allocation. On simulated experimentation, the GAPA discovers a better VM allocation (means lower energy consumption) than the baseline scheduling algorithm for solving same SVMAP. 

\section{Problem Formulation}
\label{sec:2}

\subsection{\large Terminology, notation}

We describe notation that is used in this paper as following:

	\begin{itemize}
				\item $VM_{i}$: the i-th virtual machine
				\item $M_{j}$: the j-th physical machine
				\item $ts_{i}$: start time of the $VM_{i}$
				\item $pe_{i}$: number of processing elements (e.g. cores) of the $VM_{i}$
				\item $PE_{j}$: number of processing elements (e.g. cores) of the $M_{j}$
				\item $mips_{i}$: total required MIPS (Millions Instruction Per Seconds) of the $VM_{i}$
				\item $MIPS_{j}$: total capacity MIPS (Millions Instruction Per Seconds) of the $M_{j}$
				\item $d_{i}$: duration time of the $VM_{i}$, units in seconds
				\item $P_{j}(t)$: power consumption (Watts) of a physical machine $M_{j}$
				\item $r_{j}(t)$: set of indexes of virtual machines that is allocated on the $M_{j}$ at time $t$
	\end{itemize}

\subsection{Power consumption model}

In this section, we introduce factors to model the power consumption of single physical machine. Power consumption (Watts) of a physical machine is sum of total power of all components in the machine. In \cite{Fan2007a}, they estimated power consumption of a typical server (with 2x CPU, 4x memory, 1x hard disk drive, 2x PCI slots, 1x mainboard, 1x fan) in peak power (Watts) spends on main components such as CPU (38\%), memory (17\%), hard disk drive (6\%), PCI slots (23\%), mainboard (12\%), fan (5\%). 
Some papers \cite{Fan2007a} \cite{Beloglazov2010d} \cite{Beloglazov2011} \cite{Beloglazov2012} prove that there exists a power model between power and resource utilization (e.g. CPU utilization). We assume that power consumption of a physical machine ($P(.)$) is linear relationship between power and resource utilization (e.g. CPU utilization) as  \cite{Fan2007a}\cite{Beloglazov2010d}\cite{Beloglazov2011}\cite{Beloglazov2012}. The total power consumption of a single physical server ($P(.)$) is:

\[
P(U_{cpu}) = P_{idle} + (P_{max} - P_{idle}) U_{cpu} 
\]

\[
U_{cpu}(t) = \sum_{c=1}^{PE_{j}} \sum_{i  \in  r_j(t)} \dfrac{mips_{i,c}}{MIPS_{j,c}}
\]

In which:
\begin{itemize}
%	\item $U_{cpu}$: CPU utilization of the physical machine,  $0 \leq U_{cpu} \leq 1$
	\item $U_{cpu}(t)$: CPU utilization of the physical machine at time $t$, $0 \leq U_{cpu}(t) \leq 1$
	\item $P_{idle}$: the power consumption (Watt) of the physical machine in idle, e.g. 0\% CPU utilization
	\item $P_{max}$: the maximum power consumption (Watt) of the physical machine in full load, e.g. 100\% CPU utilization 
	\item $mips_{i,c}$: requested MIPS of the $c$-th processing element (PE) of the $VM_{i}$ 
	\item $MIPS_{j,c}$: Total MIPS of the $c$-th processing element (PE) on the physical machine $M_{j}$
\end{itemize}

The number of MIPS that a virtual machine requests can be changed by its running application. Therefore, the utilization of the machine may also change over time due to application. We link the utilization with the time $t$. We re-write the total power consumption of a single physical server ($P(.)$) with $U_{cpu} (t)$ as:

\[
P(U_{cpu}(t)) = P_{idle} + (P_{max} - P_{idle}) U_{cpu}(t) 
\] 

and total energy consumption of the physical machine ($E$) in period time $ [t_0, t_1] $ is defined by:
 
\[
E = \int\limits_{t_{0}}^{t_{1}} P(U_{cpu}(t)) dt
\]

\subsection{Static Virtual Machine Allocation Problem (SVMAP)} 
\label{sec:5}

Given a set of $n$ virtual machines \{$ VM_{i}(pe_{i}, mips_{i}, ts_{i}, d_{i}) | i = 1,...,n $\} to be placed on a set of $m$ physical parallel machines \{$ M_{j}(PE_{j}, MIPS_{j}) | j = 1,...,m $\}. 
Each virtual machine $ VM_{i} $ requires $pe_{i}$ processing elements and total of $mips_{i}$ MIPS, and the $VM_{i}$ will be started at time ($ts_{i}$) and finished at time ($ts_{i} + d_{i}$) without neither preemption nor migration in its duration ($d_{i}$). We do not limit resource type on CPU. We can extend for other resource types such as memory, disk space, network bandwidth, etc.
%Resource utilization consumed by these VM are additively.

We assume that every physical machine $M_{j}$ can host any virtual machine, and its power consumption model ($P_{j}(t)$) is proportional to resource utilization at a time $t$, e.g. power consumption has a linear relationship with resource utilization (e.g. CPU utilization) \cite{Fan2007a}\cite{Ittner2007}\cite{Beloglazov2012}.

The objective scheduling is minimizing energy consumption in fulfillment of maximum requirements of $n$ VMs.

\subsection{The GAPA Algorithm}

The GAPA, which is a kind of Genetic Algorithm (GA), solves the SVMAP. The GAPA performs steps as in the Algorithm 1.

\begin{table}[t]
\centering
\begin{tabular}{p{10cm}}

Algorithm 1: GAPA Algorithm \\

\hline

\\
Start: Create an initial population randomly for $s$ chromosomes (with $s$ is population size)
\\
Fitness: Calculate evaluation value of each chromosome respectively in given population.
\\
New population: Create a new population by carrying out follows the steps:
\\
Selection: Choose the two individual parents from current population based on value of evaluation.
\\
Crossover: By using crossover probability, we create new children via modifying chromosome of parents.
\\
Mutation: With mutation probability, we will mutate at some position on chromosome.
\\
Accepting: Currently, new children will be a part of the next generation.
\\
Replace: Go to the next generation by assigning the current generation to the next generation.
\\
Test: If stop condition is satisfied then this algorithm is stopped and returns individual has the highest evaluation value. Otherwise, go to next step.
\\
Loop: Go back the Fitness step.
\\
\hline

\end{tabular} 
\end{table}

In the GAPA, we use a tree structure to encode chromosome of an individual. This structure has three levels:
\\
Level 1: Consist of a root node that does not have significant meaning.
\\
Level 2: Consist of a collection of nodes that represent set of physical machines.
\\
Level  3: Consist of a collection of nodes that represent set of virtual machines.
\\
With above representation, each instance of tree structure will show that an allocation of a collection of virtual machines onto a collection of physical machines.
The fitness function will calculate evaluation value of each chromosome as in the Algorithm 2.

\begin{table}[t]
\centering
\begin{tabular}{p{10cm}}

Algorithm 2: Construct fitness function \\

\hline

\\
powerOfDatacenter := 0 \\
For each host $\in$ collection of hosts do \\
~~~~	utilizationMips :=  host.getUtilizationOfCpu() \\
~~~~	powerOfHost :=  getPower (host, utilizationMips) \\
~~~~	powerOfDatacenter := powerOf Datacenter + powerOfHost \\
End For
\\
Evaluation value (chromosome) := 1.0 / powerOfDatacenter
\\
\hline

\end{tabular} 
\end{table}

\section{Experimental study}

\subsection{Scenarios}

We consider on resource allocation for virtual machines (VMs) in private cloud that belongs to a college or university. In a university, a private cloud is built to provide computing resource for needs in teaching and researching. In the cloud, we deploy installing software and operating system (e.g. Windows, Linux, etc.) for practicing lab hours in virtual machine images (i.e. disk images) and the virtual machine images are stored in some file servers. A user can start, stop and access VM to run their tasks. We consider three needs as following: 
\\
\begin{itemize}
\item[i] A student can start a VM to do his homework.\\
\item[ii] A lecturer can request a schedule to start a group of identical VMs for his/her students on lab hours at specified start time and in prior. The lab hours requires that the group of VMs will start on time and continue in spanning some time slots (e.g. 90 minutes).\\
\item[iii] A researcher can start a group of identical VMs to run his/her parallel application.\\
\end{itemize}

\subsection{Workload and simulated cluster}

We use workload from one-day of our university's schedule for laboratory hours on six classes in the Table \ref{tab:1}. The workload is simulated by total of 211 VMs and 100 physical machines (hosts).

We consider there are two kind of servers in our simulated virtualized datacenter, which includes two power consumption models of two power model of the IBM server x3250 (1 x [Xeon X3470 2933 MHz, 4 cores], 8GB) and another power model of the Dell Inc. PowerEdge R620 (1 x [Intel Xeon E5-2660 2.2 GHz, 16 cores], 24 GB) server with 16 cores in the Table \ref{tab:2}. The baseline scheduling algorithm (BFD), which sorts list of virtual machines in start time (i.e. earliest start time first) and using best-fit decreasing (i.e. least increased power consumption, for example MBFD \cite{Beloglazov2012}), will use four IBM servers to allocate for 16 VMs (each VM requests single processing element). Our GAPA can finds a better VM allocation (lesser energy consumption) than the minimum increase of power consumption (best-fit decrease) heuristic in our experiments. In this example, our GAPA will choose one Dell server to allocate these 16 VMs. As a result, our GAPA consumes less total energy than the BFD does.

\begin{table}
\centering
\caption{Workload of a university's one-day schedule}
\label{tab:1} 
\begin{tabular}{|c|c|c|c|c|c|c|} 
\hline Day & Subject & Class ID & Group ID & Students &  Lab. Time &  \shortstack{Duration \\(sec.)} \\
\hline
6	& 506007	& CT10QUEE	& QT01	& 5	 & ---456---------- 	& 8100	\\
6	& 501129	& CT11QUEE	& QT01	& 5	 & 123------------- 	& 8100	\\
6	& 501133	& DUTHINH6	& DT04	& 35& 123------------- 	& 8100	\\
6	& 501133	& DUTHINH5	& DT01	& 45& ---456---------- 	&  8100	\\
6	& 501133	& DUTHINH5	& DT02	& 45& ---456---------- 	& 8100	\\
6	& 501133	& DUTHINH6	& DT05	& 35& 123------------- 	&  8100	\\
6	& 501133	& DUTHINH6	& DT06	& 41& 123------------- 	&  8100	\\

\hline 
\end{tabular}
\end{table}

\subsection{Experiments}

\begin{table}[h]
\centering
\caption{Two power models of (i) the IBM server x3250 (1 x [Xeon X3470 2933 MHz, 4 cores], 8GB) \cite{SpecIBM3250} and (ii) the Dell Inc. PowerEdge R620 (1 x [Intel Xeon E5-2660 2.2 GHz, 16 cores], 24 GB) \cite{SpecDellR620} }
\label{tab:2}
\begin{tabular}{|c|c|c|c|c|c|c|c|c|c|c|c|} 
\hline Utilization	& 0.0	& 0.1	& 0.2	& 0.3	& 0.4	& 0.5	& 0.6	& 0.7	& 0.8	& 0.9	& 1 \\
\hline IBM x3250	& 41.6	& 46.7	& 52.3	& 57.9			& 65.4		& 73.0		& 80.7		& 89.5		& 99.6			& 105.0			& 113.0 \\
\hline Dell R620	  & 56.1	& 79.3	& 89.6	& 102.0		& 121.0		& 132.0		& 149.0		& 171.0		&	195.0	&	225.0	& 263.0 \\
\hline 
\end{tabular}
\end{table}

\begin{table}
\centering
\caption{Total energy consumption (KWh) of running: (i) earliest start time first with best-fit decreasing (BFD); (ii) GAPA algorithms. These GAPA use the probability mutation of 0.01 and size of population of 10. N/A means not available}
\label{tab:3} 
\begin{tabular}{|c|c|c|c|c|c|c|c|} 
\hline Algorithms	& VMs	& Hosts	& \shortstack{GA's \\Generations}  &  \shortstack{GA's Prob. \\of Crossover} &  \shortstack{Energy \\(KWh)}	& BFD/GAPA  \\

\hline BFD	& 211	& 100	& N/A & N/A & 16.858	& 1 \\

\hline  \shortstack{GAPA\_P10\_\\G500\_C25}	& 211	& 100	& 500 & 0.25 & 13.007	& 1.296 \\
\hline  \shortstack{GAPA\_P10\_\\G500\_C50}	& 211	& 100	& 500 & 0.50 & 13.007	& 1.296 \\
\hline  \shortstack{GAPA\_P10\_\\G500\_C75}	& 211	& 100	& 500 & 0.75 & 13.007	& 1.296 \\

\hline \shortstack{GAPA\_P10\_\\G1000\_C25}	& 211	& 100	& 1000 & 0.25 & 13.007	& 1.296 \\
\hline \shortstack{GAPA\_P10\_\\G1000\_C50}	& 211	& 100	& 1000 & 0.50 & 13.007	& 1.296 \\
\hline \shortstack{GAPA\_P10\_\\G1000\_C75}	& 211	& 100	& 1000 & 0.75 & 13.007	& 1.296 \\

\hline 
\end{tabular}
\end{table}

\begin{figure}[h]
\centering
\includegraphics[width=10cm, height=6cm]{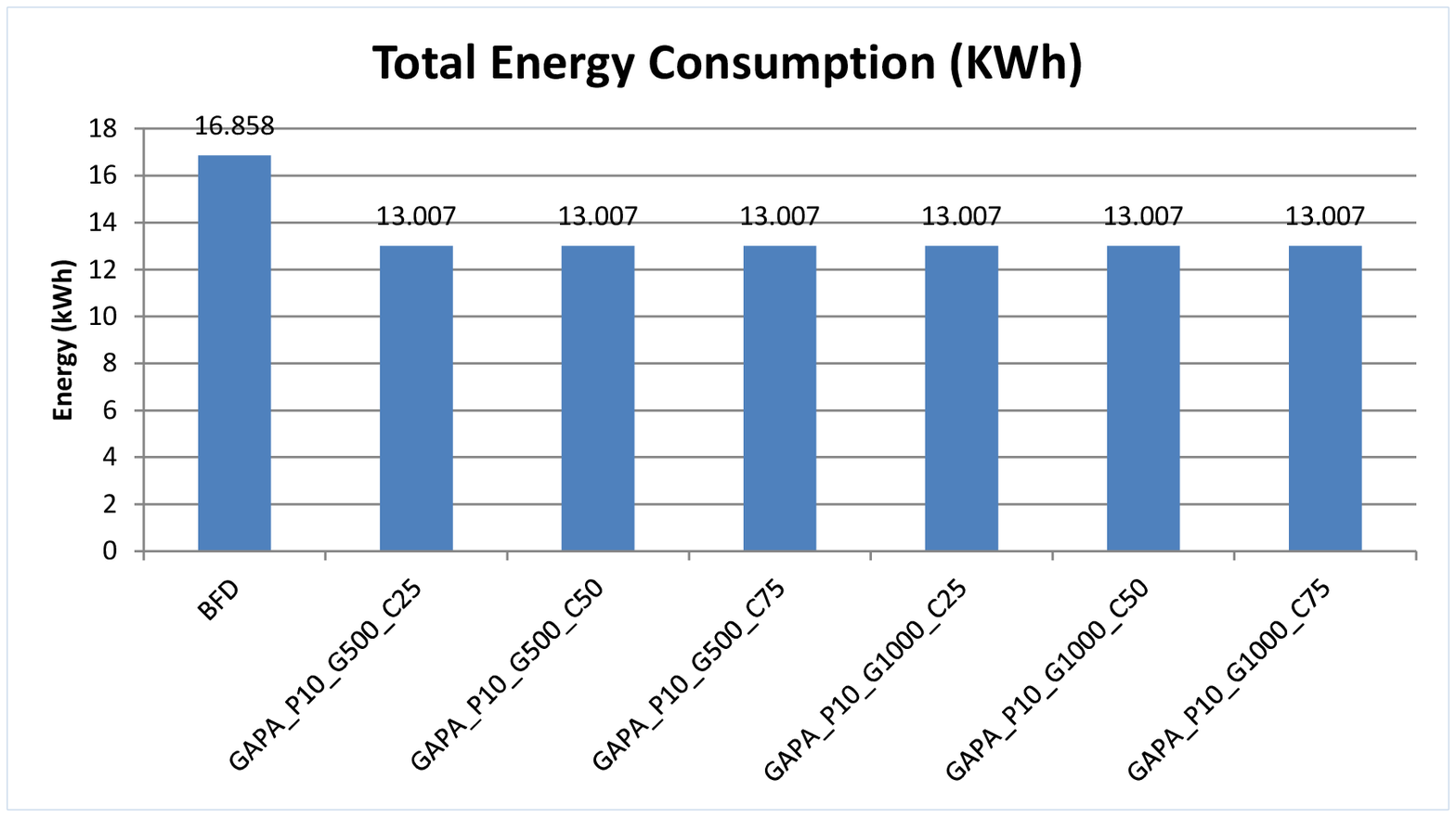}
\caption{The total energy consumption (KWh) for earliest start time first with best-fit decrease (BFD), GAPA algorithms}
%\caption{The total energy consumption (KWh) for the baseline scheduling algorithm (BFD), which is earliest start time first with best-fit decrease, and GAPA algorithms}
\label{fig:1}
\end{figure}

We show results from the experiments in the Table \ref{tab:3} and Figure \ref{fig:1}. We use a popular simulated software for a virtualized datacenter is the CloudSim \cite{Cloudsim}\cite{Beloglazov2011} to simulate our virtualized datacenter and the workload. The GAPA is a VM allocation algorithm that is developed and integrated into the CloudSim version 3.0.

On simulated experimentation, we have total energy consumptions of both the BFD and the GAPA algorithms are 16.858KWh and average of 13.007KWh respectively. We conclude that the energy consumption of the BFD algorithm is higher than the energy consumption of GAPA algorithm is approximately 130\%. In case of the GAPA, these GAPA use the probability mutation is 0.01 and size of population is 10, number of generations is \{500, 1000\}, probability of crossover is \{0.25, 0.5, 0.75\}.

\section{Related works}
\label{sec:3}
B. Sotomayor et al. \cite{Sotomayor2010} proposed a lease-based model and First-Come-First-Serve (FCFS) and backfilling algorithms to schedule best-effort, immediate and advanced reservation jobs. The FCFS and backfilling algorithms consider only performance metric (e.g. waiting time, slowdown). To maximize performance, these scheduling algorithms tend to choose free load servers (i.e. highest-ranking scores) when allocates a new lease. Therefore, a lease with single VM can be allocated on big, multi-core physical machine. This way could be waste energy, both of the FCFS and backfilling does not consider on the energy efficiency. 

S. Albers et al. \cite{Albers2010} reviewed some energy efficient algorithms which are used to minimize flow time by changing processor speed adapt to job size. G. Laszewski et al. \cite{VonLaszewski2009} proposed scheduling heuristics and to present application experience for reducing power consumption of parallel tasks in a cluster with the Dynamic Voltage Frequency Scaling (DVFS) technique. We did not use the DVFS technique to reduce energy consumption on datacenter.

Some studies \cite{Goiri2010}\cite{Beloglazov2010}\cite{Beloglazov2012} proposed algorithms to solve the virtual machine allocation in private cloud to minimize energy consumption. A. Beloglazov et al. \cite{Beloglazov2010}\cite{Beloglazov2012} presented a best-fit decreasing heuristic on VM allocation, named MBFD, and VM migration policies under adaptive thresholds. The MBFD tends to allocate a VM to such as active physical machine that would take the minimum increase of power consumption (i.e. the MBFD prefers a physical machine with minimum power increasing). However, the MBFD cannot find an optimal allocation for all VMs. In our simulation, for example, the GAPA can find a better VM allocation (lesser energy consumption) than the minimum increase of power consumption (best-fit decrease) heuristic in our experiments. In this example, our GAPA will choose one Dell server to allocate these 16 VMs. As a result, our GAPA consumes less total energy than the best-fit heuristic does.

Another study on allocation of VMs \cite{Goiri2010} developed a score-based allocation method to calculate scores matrix of allocations of $ m $ VMs to $ n $ physical machines. A score is sum of many factors such as power consumption, hardware and software fulfillment, resource requirement. These studies are only suitable for service allocation, in which each VM will execute a long running, persistent application. We consider each user job has a limited duration time. In addition to, our GAPA can find an optimal schedule for the static VM allocation problem on single objective is minimum energy consumption.

In a recently work,  J. Kolodziej et al. \cite{KolZo2012} presents evolutionary algorithms for energy management. None of these solutions solves same our SVMAP problem.

\section{Conclusions and Future works}

In a conclusion, a genetic algorithm can apply to the static virtual machine allocation problem (SVMAP) and brings benefit in minimize total energy consumption of computing servers. On simulation with workload of one-day lab hours in university, the energy consumption of the baseline scheduling algorithm (BFD) algorithm is higher than the energy consumption of GAPA algorithm is approximately 130\%. Disadvantage of the GAPA algorithm is longer computational time than the baseline scheduling algorithm.

In the future work, we concern methodology to reduce computational time of the GAPA. We also concern some other constraints, e.g. deadline of jobs. We also study on migration policies and history-based allocation algorithms.


\begin{thebibliography}{1}

\bibitem {Albers2010} Albers, S. and Fujiwara, H.: Energy-efficient algorithms. ACM Review,  Vol. 53, No. 5, (2010) pp. 86--96, doi: 10.1145/1735223.1735245

\bibitem{Ittner2007} Barroso, L.A. and H\"{o}lzle, U.: The Case for Energy-Proportional Computing, Vol. 40, pp. 33-37. ACM (2007), doi: 10.1109/MC.2007.443

\bibitem{Beloglazov2010} Beloglazov, A. and Buyya, R.: Energy Efficient Resource Management in Virtualized Cloud Data Centers, Proceedings of the 10th IEEE/ACM International Conference on Cluster, Cloud and Grid Computing, pp. 826-831. (2010) doi: 10.1109/CCGRID.2010.46

\bibitem{Beloglazov2010d} Beloglazov, A. and Buyya, R.: Adaptive Threshold-Based Approach for Energy-Efficient Consolidation of VMs in Cloud Data Centers, ACM (2010)

\bibitem{Beloglazov2012} Beloglazov, Abawajy, A., J., Buyya, R..:Energy-aware resource allocation heuristics for efficient management of data centers for Cloud computing, FGCS, Vol. 28, no. 5, pp. 755-768, (2012). DOI: 10.1016/j.future.2011.04.017

\bibitem{Beloglazov2011} Beloglazov, A. and Buyya, R.: Optimal Online Deterministic Algorithms and Adaptive Heuristics for Energy and Performance Efficient Dynamic Consolidation of Virtual Machines in Cloud Data Centers, CONCURRENCY AND COMPUTATION: PRACTICE AND EXPERIENCE, Concurrency Computat.: Pract. Exper., pp. 1-24, (2011), doi: 10.1002/cpe

\bibitem{Buyya2009a} Buyya, R., Yeo, C. S., Venugopal, S., Broberg, J., Brandic, I..: Cloud computing and emerging IT platforms: Vision, hype, and reality for delivering computing as the 5th utility, FGCS, Vol. 25, No. 6 , pp. 599-616. (2009) doi: 10.1016/j.future.2008.12.001

\bibitem{Fan2007a} Fan, X., Weber, W.-D., Barroso, L. A..: Power provisioning for a warehouse-sized computer, Proceedings of the 34th annual international symposium on Computer architecture,  pp. 13-23. ACM (2007) doi: 10.1145/1273440.1250665

\bibitem{Goiri2010} Goiri, Í., Julià, F., Nou, R., Berral, J., Guitart, J., Torres, J..:Energy-aware Scheduling in Virtualized Datacenters, in IEEE International Conference on Cluster Computing (CLUSTER 2010), (2010), pp. 58-67.

\bibitem{KolZo2012} J. Kolodziej et al. (Eds.).: A Taxonomy of Evolutionary Inspired Solutions for Energy Management in Green Computing : Problems and Resolution Methods, Advances in IntelligentModelling and Simulation, SCI 422, pp. 215--233.

\bibitem{Sotomayor2008} Sotomayor, B., Keahey, K., Foster, I..: Combining batch execution and leasing using virtual machines, Proceedings of the 17th international symposium on High performance distributed computing - HPDC '08,  pp. 87-96. ACM (2008) doi: 10.1145/1383422.1383434

\bibitem{Sotomayor2010} Sotomayor, B.: Provisioning Computational Resources Using Virtual Machines and Leases,  PhD Thesis submited to The University of Chicago, US, (2010)

\bibitem{VonLaszewski2009} Laszewski, G. V., Wang, L., Younge, A. J., He, X..: Power-aware scheduling of virtual machines in DVFS-enabled clusters, 2009 IEEE International Conference on Cluster Computing and Workshops, pp. 368--377. (2009) doi: 10.1109/CLUSTR.2009.5289182

\bibitem{Cloudsim} Calheiros, R. N., Ranjan, R., Beloglazov, A., De Rose, C. A. F., Buyya, R..: CloudSim: a toolkit for modeling and simulation of cloud computing environments and evaluation of resource provisioning algorithms, Software: Practice and Experience, vol. 41, no. 1, pp. 23--50, 2011.

\bibitem{SpecDellR620} SPECpower ssj2008 results for Dell Inc. PowerEdge R620 (Intel Xeon E5-2660, 2.2 GHz). \textit{\url{http://www.spec.org/power\_ssj2008/results/res2012q2/power\_ssj2008-20120417-00451.html}}. Last accessed: Nov. 29, 2012

\bibitem{SpecIBM3250}  SPECpower ssj2008 results for IBM x3250 (1 x [Xeon X3470 2933 MHz, 4 cores], 8GB). \textit{\url{http://www.spec.org/power\_ssj2008/results/res2009q4/power\_ssj2008-20091104-00213.html}}. Last accessed: Nov. 29, 2012

\end{thebibliography}
\end{document}